%% file: iclr2027_apple_arxiv.tex
\documentclass{apple/applemlr} 

\input{apple/apple_preamble}
\affiliation{Apple\textsuperscript{1}}
\affiliation{Caltech\textsuperscript{2}}
\author[1,2]{Rohit Dilip}
\author[1]{Tianrong Chen}
\author[1]{Yuyang Wang}
\author[2]{David Van Valen}
\author[1]{Josh Susskind}
\author[1]{Miguel Angel Bautista}

\abstract{
We introduce a new method to guide flow matching models. Our approach, which we call \textbf{\guidance}, uses the frozen internal states of an existing diffusion model to construct a guidance signal. This works using a similar principle as autoguidance, but eliminates the need for an additional forward pass at inference time and provides a reliable path to ensure that the weak and strong model share similar dynamics. We apply and benchmark this method on continuous diffusion language models, where \guidance~sets a new state-of-the-art performance on unconditional generation. When applied to a 1.7B diffusion language model, \guidance~consistently improves on multiple choice question answering benchmarks. Using our probes, we study the traditional autoguidance setting where the strong model is a weak checkpoint, and find that the weak model must come from a low-entropy region of training. These findings both provide a practical way to improve diffusion language models and shed light on the actual mechanism behind autoguidance, which is currently poorly understood. }

\metadata[Correspondence]{\sffamily Miguel Angel Bautista: \url{mbautistamartin@apple.com}; Rohit Dilip: \url{rdilip@caltech.edu}}

\date{\sffamily\today}

\usepackage{hyperref}
\usepackage{booktabs}
\usepackage{url}
\usepackage{svg}      
\usepackage{caption}
\usepackage{hyperref}
\usepackage{amsmath, amssymb}
\usepackage{url}
\usepackage{graphicx}
\usepackage{bm}
\usepackage[percent]{overpic}
\usepackage[table]{xcolor}
\usepackage{mfirstuc}
\title{How to Guide Your Language Flow}

\usepackage[most]{tcolorbox}
\usepackage{booktabs}
\usepackage{xcolor}
\definecolor{codecomment}{gray}{0.40}

\usepackage{booktabs}
\usepackage{xcolor}
\usepackage[most]{tcolorbox}

\definecolor{codegray}{HTML}{F6F6F6}
\definecolor{codebluebg}{HTML}{EEF5FC}
\definecolor{mygray}{HTML}{6B7280}

\definecolor{pyblue}{HTML}{2F6F9F}
\definecolor{pypurple}{HTML}{8A5A9B}
\definecolor{pycomment}{HTML}{6A737D}

\newcommand{\pyfunc}[1]{\textcolor{pyblue}{#1}}
\newcommand{\pyconst}[1]{\textcolor{pypurple}{#1}}
\newcommand{\pycomment}[1]{\textcolor{pycomment}{#1}}

\newtcolorbox{pseudocodeblock}[1]{
    colback=#1,
    colframe=#1,
    boxrule=0pt,
    arc=1.5pt,
    left=4pt,
    right=4pt,
    top=3pt,
    bottom=3pt,
    boxsep=0pt,
    before skip=2.5pt,
    after skip=2.5pt,
}

\definecolor{findingblue}{HTML}{F1F6F9}
\definecolor{findingborder}{HTML}{B7CBD6}

\usepackage{fontawesome5}

\usepackage[most]{tcolorbox}
\usepackage{xcolor}

\usepackage[most]{tcolorbox}
\usepackage{xcolor}

\definecolor{codegray}{HTML}{F5F6F7}
\definecolor{codeblue}{HTML}{EEF4FA}
\definecolor{myblue}{HTML}{2A5C8D}
\definecolor{mygreen}{HTML}{2E7F72}
\definecolor{myorange}{HTML}{C9743D}
\newtcolorbox{pseudocodebox}[2][]{%
    enhanced,
    colback=#2,
    colframe=#2,
    boxrule=0pt,
    arc=2pt,
    left=5pt,
    right=5pt,
    top=4pt,
    bottom=4pt,
    boxsep=0pt,
    before skip=3pt,
    after skip=3pt,
    #1
}

\definecolor{findingblue}{HTML}{E8F1FA}
\definecolor{findingborder}{HTML}{52789A}

\usepackage{booktabs}
\usepackage{xcolor}

\definecolor{codecomment}{gray}{0.40}
\newtcolorbox{findings}{
    enhanced,
    colback=findingblue,
    colframe=findingborder,
    boxrule=1.0pt,
    arc=2.5mm,
    outer arc=2.5mm,
    left=3.5mm,
    right=3.5mm,
    top=2.5mm,
    bottom=2.5mm,
    boxsep=0pt,
    before skip=8pt,
    after skip=8pt,
    before upper={
        \textcolor{findingborder}{\normalsize\faBookmark}
        \hspace{1.5mm}
        \textbf{Finding.}\ 
    },
}

\usepackage[most]{tcolorbox}
\usepackage{xcolor}

\definecolor{samplegray}{HTML}{F3F4F6}
\definecolor{sampleblue}{HTML}{EAF2F8}

\definecolor{sampleheader}{HTML}{6B7280}
\usepackage[most]{tcolorbox}
\usepackage{xcolor}

\definecolor{samplegray}{HTML}{F7F7F7}
\definecolor{sampleblue}{HTML}{F2F6FA}

\definecolor{sampleborder}{HTML}{A6A6A6}
\definecolor{sampleheader}{HTML}{555555}
\usepackage[most]{tcolorbox}
\usepackage{xcolor}

\definecolor{samplegray}{HTML}{F7F7F7}
\definecolor{sampleblue}{HTML}{F2F6FA}
\usepackage[most]{tcolorbox}
\usepackage{xcolor}

\definecolor{samplegray}{HTML}{F7F7F7}
\definecolor{sampleblue}{HTML}{F1F6FA}
\definecolor{sampleborder}{HTML}{A0A0A0}
\definecolor{sampleheader}{HTML}{555555}

\usepackage[most]{tcolorbox}
\usepackage{xcolor}

\definecolor{samplegray}{HTML}{F7F7F7}
\definecolor{sampleblue}{HTML}{EEF4FA}
\definecolor{sampleborder}{HTML}{999999}
\usepackage[most]{tcolorbox}
\usepackage{xcolor}

\definecolor{samplegray}{HTML}{F7F7F7}
\definecolor{sampleblue}{HTML}{EEF4FA}
\definecolor{sampleborder}{HTML}{999999}

\newenvironment{modelsample}[2][samplegray]
{%
  \begin{tcolorbox}[
    enhanced,
    breakable,
    colback=#1,
    colframe=sampleborder,
    boxrule=0.6pt,
    arc=1.5mm,
    left=3.5mm,
    right=3.5mm,
    top=3mm,
    bottom=3mm,
    before skip=8pt,
    after skip=8pt
  ]
  {\small\sffamily\bfseries #2\par}
  \vspace{1.5mm}
  {\color{sampleborder}\hrule height 0.5pt}
  \vspace{2mm}
  \small\sffamily
}
{%
  \end{tcolorbox}
}

\usepackage{xcolor}

\newcommand{\x}{\mathbf{x}}
\newcommand{\vel}{\mathbf{v}}
\newcommand{\noise}{\bm{\epsilon}}

\usepackage{xcolor}

\usepackage{xcolor}

\definecolor{rohitcolor}{HTML}{3B6EA8}
\definecolor{yuyangcolor}{HTML}{8E5AA7}
\definecolor{miguelcolor}{HTML}{3F8C78}
\definecolor{tianrongcolor}{HTML}{C06C4E}
\definecolor{joshcolor}{HTML}{B58A32}

\newcommand{\guidance}{probe guidance}
\newcommand{\Guidance}{Probe guidance}

\begin{document}

\maketitle

\applefootnote{{\sffamily Work done while R.D. was an intern at Apple.}}

\section{Introduction}
Diffusion and flow-matching models are the dominant paradigm for generation in continuous space, from images to videos to proteins~\citep{esser2024scaling, ho2022video, lai2025principles, wang2026simplefold, abramson2024accurate}. Their success has inspired efforts to train diffusion language models (dLMs), which apply the principles of diffusion to text generation~\citep{li2025survey, shi2024simplified, sahoo2024simple} in discrete spaces. These approaches contrast with typical autoregressive models~\citep{radford2019language} by generating tokens simultaneously from a corrupted state, as opposed to one at a time in sequential order. While most language diffusion models are formulated over discrete space, a recent line of work studies \emph{continuous} diffusion models that define an interpolant between a Gaussian prior and data as a continuous random variable~\citep{chen2026langflow, lee2026flow, davis2026scaling, roos2026categorical, hu2026elf}.

These models inherit many of the advances in continuous diffusion. A particularly powerful technique is guidance, where the output of a diffusion model is adjusted at inference-time for better alignment with a given reward. The most widespread guidance method is classifier-free guidance (CFG), where the guiding signal comes from the difference between a class-conditioned model and an unconditional model~\citep{ho2022classifier, chung2025cfg++}. More recent work introduced autoguidance, where the guiding signal comes from the difference between a primary model and weaker model~\citep{karras2024guiding}. 






\begin{figure}[t]
    \centering

    \includegraphics[width=\linewidth]{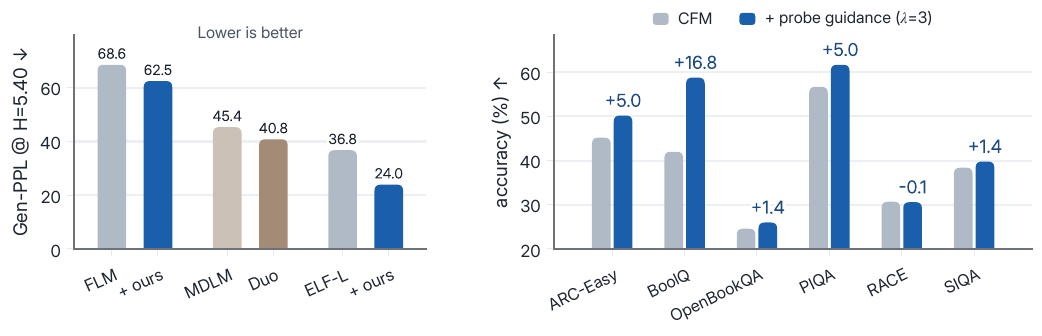}

    \vspace{2pt}

    \begin{minipage}[t]{0.49\linewidth}
        \centering
        \small
        \textbf{(a)} Unconditional generation.
    \end{minipage}
    \hfill
    \begin{minipage}[t]{0.49\linewidth}
        \centering
        \small
        \textbf{(b)} Multiple choice evaluation.
    \end{minipage}

    \caption{\Guidance~improves both data space (FLM) and latent space (ELF) models on (a) unconditional generation and (b) multiple choice evaluation. On unconditional generation, \guidance~takes ELF-L down to a genPPL score of 23 at the OWT dataset entropy 5.40. In multiple choice evaluations, we find improvement or parity across categories.}
    \label{fig:base}
\end{figure}

Despite the importance of guidance in diffusion, it has been challenging to apply to dLMs. Some dLMs like ELF~\citep{hu2026elf} use CFG with self-conditioning, where the conditioning signal is the model's own estimation of the posterior mean. This provides a knob to traverse the quality diversity tradeoff. However, although self-conditioning significantly improves dLM quality~\citep{chen2026langflow}, the reasons why it works are poorly understood; as we show in our work, \guidance~consistently Pareto dominates self-conditioning. Moreover, because self-conditioning requires an additional unconditional pass, it adds training cost and makes scaling harder. FLM~\citep{lee2026flow} uses autoguidance with a weak model obtained via isotropic dropout on the strong model; although this slightly improves performance, the weak model does not share the strong model dynamics, which is known to be important in autoguidance~\citep{karras2024guiding}. In both cases, sweeping through guidance strength traces out a Pareto curve between quality and diversity.

This work introduces a new guidance method, which we call \guidance, and studies its effects in the diffusion language model setting. The guidance signal is obtained from a lightweight MLP probe ($\approx 2\%$ of model compute) trained on frozen hidden states inside a dLM. During inference, these probes are used to guide unconditional generation. In this work, we investigate the following settings:

\textbf{Efficiency} Compared to autoguidance, which requires a full additional pass of a weaker model, \guidance~requires $1.4-2.0\times$ fewer FLOPs. The probes can be trained using $\approx 1\%$ additional training compute.

\textbf{Unconditional generation:} Current continuous dLMs use CFG with self-conditioning or autoguidance to guide generations. Applied to latent models like ELF, \guidance~achieves a state-of-the-art GenPPL score of 23 at entropy 5.40. On data space models like FLM, \guidance~dominates the autoguided Pareto frontier.

\textbf{Multiple choice question answering: }\Guidance~consistently improves the performance of a 1.7B continuous dLM~\citep{davis2026scaling} across a suite of likelihood evaluations. As a supplementary contribution, we show how to adapt the likelihood estimator to make it amenable to guidance.

\textbf{Guidance dynamics: }Autoguidance has been challenging to achieve in language. We study the training dynamics of dLMs and find that both latent and data space dLM training show a period of rapid entropy increase. We demonstrate that autoguidance requires the weak model to come from the low-entropy region before the climbout. 

Practically, \guidance~provides an efficient and reliable way to improve continuous diffusion language model performance. More generally, the use of frozen features provides provides a simple way to control the strength of the weak model and study the requirements for effective guidance.

\section{Background and related work}

\textbf{Diffusion language models.} Diffusion and flow matching models push noise to data by parameterizing an ODE or SDE through a neural network. Clean data is corrupted at varying levels of noise, then a network is trained to estimate the clean data conditioned on the current level of corruption and the noised data. Early work~\citep{ho2020denoising} formulated this process as a sequence of discrete transitions, while subsequent work on score matching~\citep{song2020score} and flow matching~\citep{lipman2022flow} modeled the network as a continuous field between distributions. Explicitly, flow matching constructs a conditional interpolant by sampling $\noise\sim p_\text{prior}$ and $\mathbf{x}\sim p_\text{data}$ and writing $\x_t = \alpha_t \x + \sigma_t \noise$, where the standard choice is the linear interpolant $\alpha_t=t, \sigma_t=1-t$. Given a neural parameterization of the velocity $\vel=\dot{\x} = \x - \noise$, one can sample from $p_\text{data}$ by sampling $\x_0\sim p_\text{prior}$ and running the ODE $d\x = \vel\,dt$. A full intro to flow matching is in Appendix~\ref{sec:app:flow_matching}. We follow recent work that parameterizes the velocity through the posterior mean~\citep{li2026back}. 

\begin{equation}
    \label{eq:x_to_vel}
    \vel = \frac{\mathbb{E}[\x|\x_t, t] - \x_t}{1-t}= \frac{\x_\theta(\x_t, t) - \x_t}{1-t}
\end{equation}

This formulation can be optimized using any Bregman divergence, which is particularly helpful in the discrete setting (see Appendix~\ref{sec:app:math}).

\textbf{Diffusion language models} Diffusion models were generalized to discrete space using discrete state transitions~\citep{austin2021structured}. More recently, a number of methods have considered how continuous formulations like flow matching can be applied in the discrete setting. Embedding space methods like ELF, Diffusion-LM~\citep{li2022diffusion}, and CDCD~\citep{dieleman2022continuous} encode high dimensional vocabularies in compressed embeddings and perform diffusion in the embedding space. Simplex models like TESS~\citep{mahabadi2024tess} and Bit-diffusion~\citep{chen2022analog} form diffusion trajectories directly on the probability simplex. 

\textbf{Guidance}
Guidance can be broadly expressed in the following form, where the particular form of guidance is determined by the guidance vector $D(\mathbf{x}_t, t, c)$.
\begin{equation}
    \vel_\text{guided}(\mathbf{x}_t, t, c) = \vel_\theta(\mathbf{x}_t, t, c) + \lambda \,\mathbf{D}(\mathbf{x}_t, t, c)
\end{equation}

For example, classifier-free guidance~\citep{ho2022classifier} sets $D(\mathbf{x}_t, t, c) = v_\theta(\mathbf{x_t}, t, c) - v_\theta(\mathbf{x_t}, t, \varnothing)$. Autoguidance~\citep{karras2024guiding} allows for unconditional sampling by setting $\mathbf{D}(\mathbf{x}_t, t, c) = \vel_\text{strong}(\mathbf{x}_t, t, c) - \vel_\text{weak}(\mathbf{x}_t, t, c)$, where $\vel_\text{strong}$ and $\vel_\text{weak}$ are strong and weak models. Our work is inspired by autoguidance, which requires that the strong and weak models have correlated dynamics (e.g., isotropically corrupting the model is insufficient to produce a good guidance signal). For this reason, autoguidance is commonly implemented using an early checkpoint of a main model, which fulfills this requirement but doubles the FLOPs at inference time. We discuss similar guidance works in Appendix~\ref{sec:app:similar_guidance}.

\begin{figure}[t]
    \centering

    \begin{minipage}[t]{0.54\linewidth}
        \vspace{0pt}
        {\scriptsize(a)}
        \vspace{1mm}

        \centering
        \includegraphics[width=\linewidth]{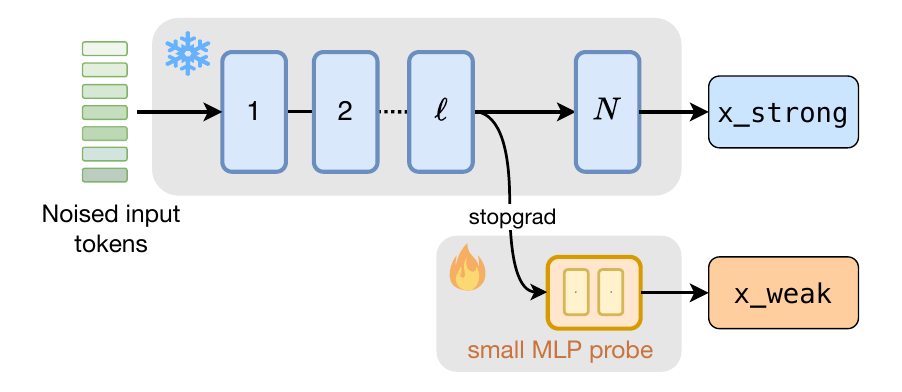}
    \end{minipage}
    \hfill
    \begin{minipage}[t]{0.43\linewidth}
        \vspace{0pt}
        \hspace{-3.5mm}{\scriptsize(b)}
        \vspace{3mm}

        \footnotesize
        \ttfamily
        \renewcommand{\arraystretch}{0.95}

        \begin{pseudocodeblock}{codegray}
        \begin{tabular}{@{}l@{\;}c@{\;}l@{}}
            \multicolumn{3}{@{}l}{
                \pycomment{Classifier-free guidance}
            }\\[-0.5mm]
            \midrule
            D = \pyfunc{net}(z\_t, c) - \pyfunc{net}(z\_t, \pyconst{0})
        \end{tabular}
        \end{pseudocodeblock}

        \begin{pseudocodeblock}{codegray}
        \begin{tabular}{@{}l@{\;}c@{\;}l@{}}
            \multicolumn{3}{@{}l}{
                \pycomment{Autoguidance}
            }\\[-0.5mm]
            \midrule
            D = \pyfunc{net\_strong}(z\_t, c) - \pyfunc{net\_weak}(z\_t, c)
        \end{tabular}
        \end{pseudocodeblock}

        \begin{pseudocodeblock}{codebluebg}
        \begin{tabular}{@{}l@{\;}c@{\;}l@{}}
            \multicolumn{3}{@{}l}{
                \pycomment{Probe guidance}
            }\\[-0.5mm]
            \midrule
            \textbf{\color{myblue}x\_{strong}}, \textbf{\color{myorange}h\_l} = \pyfunc{net}(z\_t, c) \\
            D = \textbf{\color{myblue}x\_{strong}} - \textbf{\color{myorange} {probe}(h\_l)}
        \end{tabular}
        \end{pseudocodeblock}

    \end{minipage}

    \caption{
        Overview of our method.
        (a) To train our probes, we detach hidden states from the frozen main model trunk and use an MLP to predict the flow target. This can be done during training, or with a short fine-tuning stage on a frozen backbone. (b) Classifier free guidance and autoguidance require two separate evaluations, whereas \guidance~only requires an evaluation of a small MLP on top of hidden states computed during the forward pass.
    }
    \label{fig:fig1_overview}
\end{figure}


\section{Method}
\textbf{Guidance using frozen states.} Given a pretrained dLM, we extract hidden states $\mathbf{h}_\ell$, where $\ell$ indexes the layer. For a flow model trained with divergence $\mathcal{D}$ (e.g., a cross-entropy loss for FLM or an MSE loss for ELF), we train the probe using the following objective:

\begin{equation}
    \mathcal{L} = \mathbb{E}_{\x, t, \noise}\left[\sum_{\ell}\mathcal{D}\left(\x_\theta^w(\x_t, t, \ell), \mathbf{x}\right)\right]\qquad \x_\theta^w(\x_t, t) = \text{MLP}[\text{sg}(\mathbf{h}_\ell)]
\end{equation}

where $\mathbf{x}$ is clean data. In other words, we train our probes (i.e., a weak version of the model) using the hidden states from the main model trunk. The stop-gradient removes the need for balancing multiple losses and prevents the weak path from corrupting the training dynamics of the main trunk. It also allows us to simultaneously train multiple probes on the same model trunk. During inference, we integrate using the guided velocity. Explicitly, let $\x^{\{s,w\}}_\theta$ represent the strong ($s$) and weak ($w$) outputs of the model. Then 

\begin{equation}
    \x_\text{guided} = \x^s_\theta(\x_t, t) + (\lambda - 1)\left(\x^s_\theta(\x_t, t) - \x^w_\theta(\x_t, t)\right)
\end{equation}

and the velocity update is obtained using Equation~\ref{eq:x_to_vel}. We adopt the convention in which $\lambda=1$ recovers the unguided strong model and $\lambda>1$ extrapolates along the guidance direction $\x^s_\theta-\x^w_\theta$; Figure~\ref{fig:fig1_overview} shows our training setup and compares the inference cost of \guidance~against CFG and autoguidance.

\textbf{Training and inference} Because training the probes does not affect the gradients of the main model trunk, the cost of training our~\guidance~model is the gradient-free cost of running the model trunk up to the exit layer, plus the attention-free cost of training an MLP probe. As a result, the cost of training these probes measured via the number of training steps and the cost per training step is minimal. At inference time, the base hidden state is always computed for the primary model output, so the extra inference cost is only the additional MLP pass. This contrasts with standard autoguidance, which requires two forward passes: one for the strong model and one for the weak model (which is usually an equivalent compute version of the strong model). All of our experiments do prediction in $\x$ space and convert to velocity. 

\section{Unconditional~\guidance~experiments}
\label{sec:ablations}
In this section, we demonstrate how \guidance~can improve unconditional generation across both latent and data space language diffusion models. We consider two main representative methods: ELF, a latent space model trained in a T5-latent space and FLM, a data space model that operates directly on raw tokens~\citep{hu2026elf, lee2026flow}. In both cases, we use released checkpoints to train probes and mirror the inference settings described in the main paper (see  Appendix~\ref{sec:app:experiments}).

\begin{figure}
    \centering
    \includegraphics[width=\linewidth]{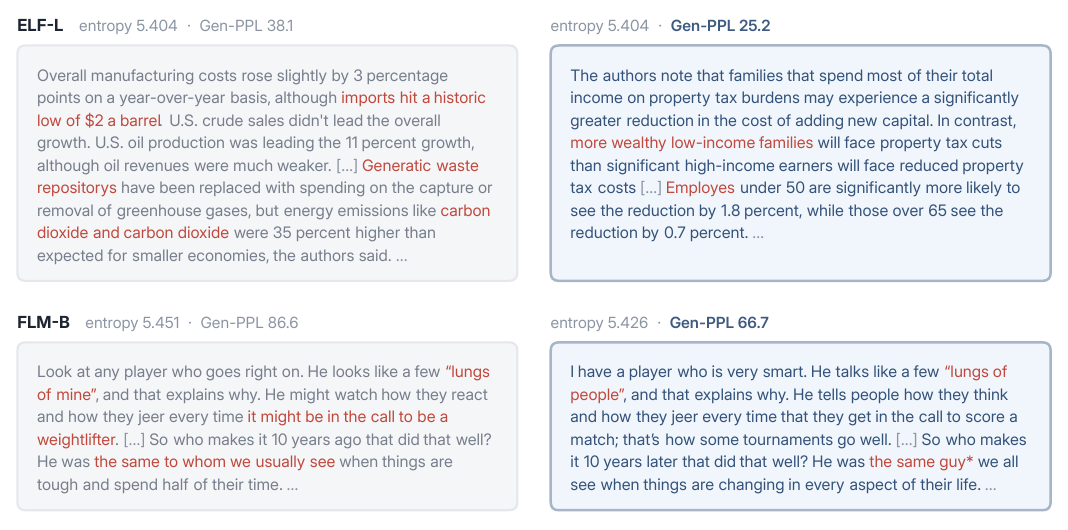}
    \caption{Real text examples using ELF-L and FLM-B. Unguided on left and with \guidance~on right. Errors highlighted in red. Guided ELF-L is notably more fluent than unguided. While FLM-B is a small model, so both guided and unguided models do not read as fluent English, it has fewer spelling mistakes and better sentence constructions (e.g., ``the same to whom we usually see'' $\rightarrow$ ``the same guy''.}
    \label{fig:real_text}
\end{figure}
\textbf{Datasets and evaluations} We measure unconditional generation using generative perplexity (GenPPL) and unigram entropy. GenPPL computes the perplexity of generated sequences under a GPT2-L model (lower GenPPL means higher quality). Unigram entropy computes the entropy of the empirical distribution implied by the token counts. While prior works often report a single GenPPL value, ~\citep{pynadath2026generative} points out that these metrics are often misleading, and only correspond to dominance in a restricted part of the phase space. To resolve these ambiguities, we plot the full Pareto curve to compare different metrics. To trace the Pareto curve, we vary self-conditioning strength for ELF. For FLM, we follow the prescription in ~\cite{lee2026flow} where we guide the main model using a weaker model obtained by applying dropout to the weights. We perform all experiments using the OpenWebText (OWT) dataset and report GenPPL/entropy by generating 1000 samples.

\textbf{Training and inference.} Our probe training mimics each respective model's training. For ELF, we use a logit-normal noise schedule with mean -1.5 and CFG strength sampled log-uniformly over [0.5, 5.0], and we generate samples using the recommended 32 or 64 step ``SDE"\footnote{Not a true SDE, it has different marginals than the ODE} sampler. In our ~\guidance~curves, we set the self-conditioning CFG strength to 1.0 (no self-conditioning guidance) and sweep through the guidance strength. For FLM, we use the released -B model, which contains 170M parameters over the GPT-2 vocabulary (50k tokens). Due to the large number of parameters in the unembedding head, we share the unembedding head with the probe and backpropagate through without updating parameters. For both ELF and FLM, we train probes with the AdamW optimizer with learning rate 0.0003 and batch size 256.

\subsection{Ablations}
In this section, we systematically explore the design space of \guidance. We center our investigation around the following axes: the probe position along the trunk (i.e. the depth at which the probe is attached), the connector between the main trunk and the probe, the compute in the probe itself, and the number of training FLOPs used to train the probe. Fundamentally, all of these axes are different ways of controlling how closely the strong and weak model align, but they each come with their own biases (for instance, a weak probe at the penultimate layer will be better than a very strong probe at an early layer).

\begin{figure}
    \centering
    \includegraphics[width=\linewidth]{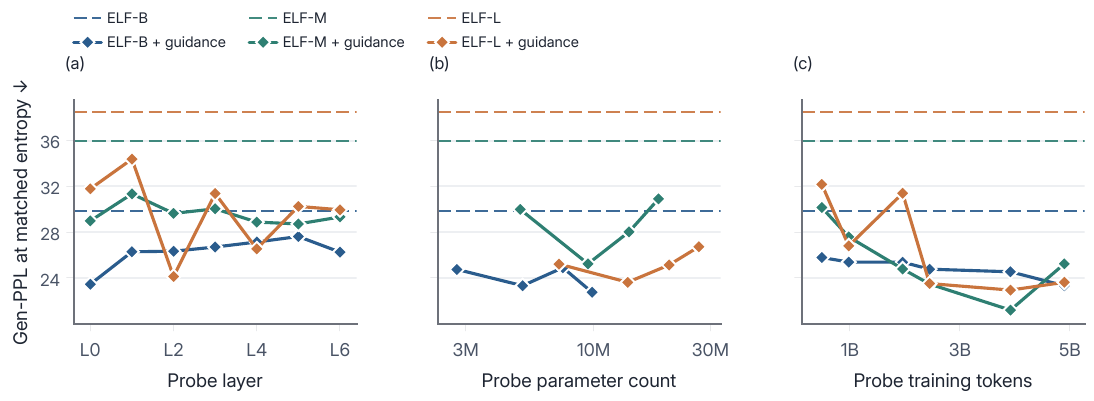}
    \caption{\textbf{Probe architecture ablations.} We study (a) which layer to attach the probe, (b) the strength of the probe in FLOPs, and (c) the amount of training per probe. We consider GenPPL at entropy $H=5.2,5.3,5.4$ for ELF {\color{myblue} \textbf{B}}, {\color{mygreen} \textbf{M}}, and {\color{myorange} \textbf{L}} respectively. Unguided performance is dashed for each model size. We find remarkably steady performance across model sizes. All configurations beat the unguided baseline (dashed). Relatively small probes are still effective, and once the probe layer and size are fixed we find that performance is quite steady with respect to the amount of training. Unguided model baselines are dashed, FLM results in Appendix~\ref{sec:app:flm-ablations}}
    \label{fig:ablations}
\end{figure}

  \begin{figure}[t]
      \centering
      \begin{minipage}[c]{0.42\linewidth}
          \centering\includegraphics[width=\linewidth]{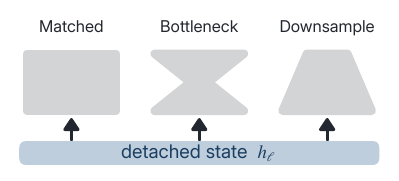}
      \end{minipage}\hfill
      \begin{minipage}[c]{0.2\linewidth}
          \centering\small
          \textbf{ELF-L}\\[2pt]
          \begin{tabular}{@{}lr@{}}
            \toprule
            Method & Gen-PPL\\
            \midrule
            \rowcolor{gray!15}
            \textit{Baseline} & \textit{36.8} \\
            \midrule
            Downsample    & \textbf{23.2} \\
            Matched width & \textbf{24.5} \\
            Bottleneck    & 30.7 \\
            \bottomrule
        \end{tabular}
      \end{minipage}\hfill
      \begin{minipage}[c]{0.25\linewidth}
          \centering\small
          \textbf{FLM-B}\\[2pt]
           \begin{tabular}{@{}lr@{}}
            \toprule
            Method & Gen-PPL \\
            \midrule
            \rowcolor{gray!15}
            \textit{Baseline} & \textit{79.3} \\
            \midrule
            Downsample    & 85.9 \\
            Matched width & \textbf{65.2} \\
            Bottleneck    & 71.1 \\
            \bottomrule
        \end{tabular}
      \end{minipage}
      \caption{(Left) Schematic of our three connector designs. Tables report Gen-PPL at H=5.40 for ELF (center) and FLM (right).}
      \label{fig:bottlenecks}
  \end{figure}

\textbf{Probe positioning:} We first study the impact of probe positioning. This is important because early layers typically provide little contextual information, while late layers provide representations that already encode the main model trunk's prediction, so the weak and strong models collapse and the guidance direction vanishes. We find that placing the probes at early layers consistently leads to the best performance (measured by GenPPL at fixed entropy). This has two implications. First, it reduces the cost of training the probes, because the dominant cost (the main model forward pass) is restricted to the first few layers. Second, it suggests that a good weak model for guidance may require very little computation, which contrasts with current approaches where the weak model is equally expensive as the strong model~\citep{karras2024guiding}.


\textbf{Trunk-probe connector: } We next study the transition between the trunk and the probe. Inspired by works that add a bottleneck layer akin to classical manifold learning~\citep{li2026back,hinton2006reducing,vincent2008extracting}, we consider three choices: matching the probe width to the model width, setting the probe width to the model width with an initial bottleneck layer, and setting the probe width \emph{smaller} than the model width. These are depicted schematically in Figure~\ref{fig:bottlenecks}. 

While all three designs work well and outperform the base model, we find dynamics differ between ELF and FLM. Latent space models like ELF can benefit from aggressive downsampling, which we attribute in part to the fact that the weak model does not need to be particularly good, so a small amount of additional computation suffices. FLM, however, clearly benefits from matching widths. For this reason, we default to matched width and discuss further in Appendix~\ref{sec:app:connector_choice}.

\textbf{Probe FLOPs: }Once the bottleneck is fixed, how should compute within the probe be allocated? Because our probes are MLPs and do not mix information across tokens, parameter count is directly proportionate to FLOPs. Figure ~\ref{fig:ablations}b shows these sweeps. We consistently find that a small number of FLOPs is sufficient, and the optimal number of FLOPs is flat between three and seven percent of main model trunk compute. While we expect that the optimal probe design will be idiosyncratic to data and model architectures, the stability of these results is reassuring; it suggests that a small probe can provide a useful guidance signal.

\textbf{Probe checkpoint: }Finally, we study how the probe checkpoint evolves during training. We sweep guidance at different checkpoints and consistently observe a minimum after a few thousand steps.  We find that training probes for $\sim10^9$ tokens (about $10^4$ steps on 8 H100 GPUs) suffices. By comparison, training the dLM trunk is typically $\sim10^{11}$ or more tokens. Practically speaking, the relatively low number of steps required for good guidance is a desirable quality. Moreover, for a fixed probe position, we do not need to optimize the checkpoint; training longer does not significantly harm performance.

\begin{findings}
Probes should be placed at early layers, and trained for a short run of a few billion tokens over a few thousand steps. The probes themselves only need a small fraction of total model compute.
\end{findings}

\begin{figure}
    \centering
    \includegraphics[width=\linewidth]{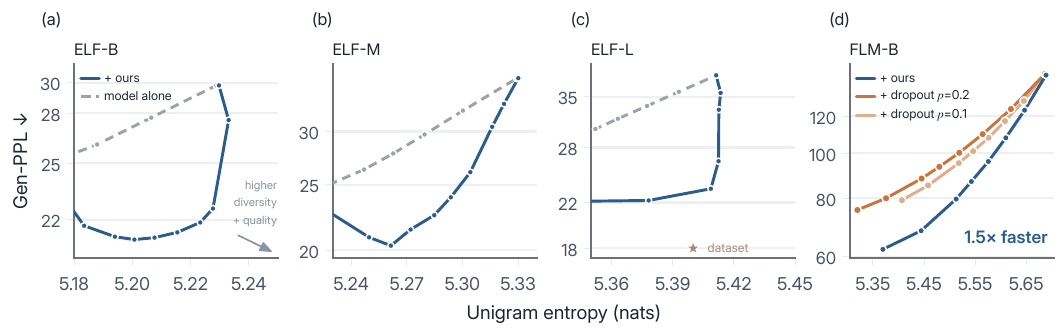}
    \caption{Across latent (a,b,c) and data space (d) released models, probe guidance consistently improves the generation quality gap. On ELF, each point represents a different CFG strength; on FLM, each point is a different autoguidance strength using a weak model obtained via dropout. \Guidance~consistently improves over the base performance. On FLM (d), probe guidance is $\approx1.5\times$ cheaper than the prescribed dropout autoguidance and improves performance by nearly 20 genPPL points (see Appendix~\ref{sec:app:speedup} for discussion).}
    \label{fig:pareto_curve_results}
\end{figure}

\textbf{\Guidance~establishes a new Pareto frontier.} Integrating these results reveals that~\guidance~consistently establishes a new Pareto frontier across model sizes and model types, meaning the probe-guided models achieve higher quality for the same level of diversity.  Pareto curves expressing the quality-entropy tradeoff are shown in Figure~\ref{fig:pareto_curve_results}. At $H=5.40$ (the approximate entropy of the OWT dataset), \guidance~improves GenPPL from $38\rightarrow 23$ on ELF-L. The FLM-B base model improves Gen-PPL by 16 points with a lower inference overhead compared to dropout autoguidance. We show sample text in Figure~\ref{fig:real_text}, with further examples in Appendix~\ref{sec:app:samples}. At the same entropy, ELF-L is noticeably clearer with fewer spelling mistakes and grammatical errors.

\section{\Guidance~improves Multiple Choice Question Answering}

We next evaluate the effect of \guidance~on multiple choice question answering (MCQA), where the objective is to measure the zero-shot performance of a model on multiple choice questions. MCQA is important both as an evaluation metric for large models and as a step beyond simple unconditional generation. We build on ~\citep{davis2026scaling, roos2026categorical}, which scales a class of continuous dLMs called Categorical Flow Models (CFMs) to 1.7B parameters and trains on the Nemotron pretraining dataset~\citep{su2025nemotron}. We train probes using our prescription in Section~\ref{sec:ablations}. Figure~\ref{fig:jump_uncond_amort}a shows unconditional performance, where we develop the CFM Pareto curve by using dropout autoguidance. As with FLM, \guidance~Pareto dominates the base performance without incurring additional FLOP costs. 

\begin{figure}
    \centering
    \includegraphics[width=\linewidth]{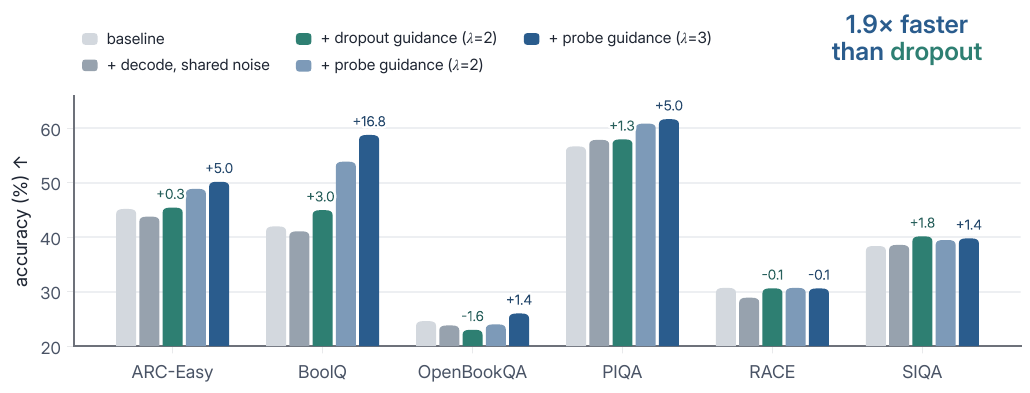}
    \caption{Performance on MCQA across benchmarks. We show the baseline reproduced performance of the flow matching model, performance with shared noise, and performance with guidance. Guidance improves MCQA across all benchmarks and closes the gap between continuous and discrete dLMs. As suggested in FLM, we include guidance based on dropout of the base model, which generally underperforms~\guidance~and costs twice the FLOPs. Tabular results in Table~\ref{tab:mcqa_full}.}
    \label{fig:mcqa}
\end{figure}

Because diffusion models do not natively provide a tractable likelihood, evaluating zero-shot performance requires a likelihood proxy. For an interpolant $I_t = \alpha_t x_1 + (1 - \alpha_t) \epsilon, \epsilon\sim\mathcal{N}(0, 1)$, \citep{davis2026scaling} suggests the following proxy

\begin{equation}
\label{eq:theory_bound}
  -\log p_\theta(x)\;\le\;
  \mathbb{E}_{t\sim U[0,1],\,x_0\sim\mathcal
  N(0,I)}\!\left[{\frac{2\dot\alpha_t\,\alpha_t}{(1-\alpha_t)^3}}\,\ell_{\mathrm{CE}}\big(e_x,\,\pi_\theta(x_t,t)\big)\right]
  \;+\;\ell_{\mathrm{CE}}\big(x,\,\pi_\theta(e_x,1)\big)
\end{equation}

This provides a certified bound; the first term is the signal-to-noise weighted time integral of the denoising cross entropy, and the second term accounts for the final readout. In practice, the weighting of the first term is unstable due to the $1/(1-\alpha_t)$ factor; \citep{davis2026scaling} therefore uses the following unweighted KL bound

\begin{equation}
-\log p_\theta(x)\;\le\;\mathbb{E}_{t\sim U[0,1],\,x_0}\!\big[\,\ell_{\mathrm{CE}}(e_x,\pi_\theta(x_t,t))\big]
\end{equation}

The second term vanishes for the true denoiser (see Appendix~\ref{sec:app:kl_proof}). Guidance shifts away from the trained distribution, which amplifies errors in the final decoding step. For this reason, we construct a new guided scoring mechanism in Equation~\ref{eq:score} which maintains the unweighted cross entropy integral but keeps the final decoding term. See Appendix~\ref{sec:app:kl_proof} for discussion.

\begin{equation}
\label{eq:score}
       S_\theta(x)\;=\;-\,\mathbb{E}_{t\sim U[0,1],\,x_0\sim\mathcal N(0,I)}\!\Big[\textstyle\ell_{\mathrm{CE}}\big(x_j,\,\tilde\pi_\theta(x_t,t)\big)\Big]-\ell_{\mathrm{CE}}\big(x_j,\,\tilde\pi_\theta(e_x,1)\big)
\end{equation}

Here, $\tilde\pi$ represents the \emph{guided} probability estimation. We evaluate the guided model using our new estimator on a standard suite of multiple choice question answering tasks across reasoning and reading comprehension: ARC-e~\citep{clark2018think}, BoolQ~\citep{clark2019boolq}, SIQA~\citep{zadeh2019social}, RACE~\citep{lai2017race}, OBQA~\citep{mihaylov2018can}, and PIQA~\citep{bisk2020piqa}, consistent with other large diffusion model methods~\citep{davis2026scaling, sahoo2026scaling}. In Figure~\ref{fig:mcqa}, we measure the effect of~\guidance~on these metrics, as well as the new scoring estimator that we introduce. Adding guidance consistently improves all metrics, including +16.8 points on BoolQ and +1.4 on the challenging SIQA benchmark. The addition of guidance brings CFM close to parity with leading discrete diffusion models like Duo and MDLM, see Table~\ref{tab:mcqa_full}~\citep{sahoo2024simple, sahoo2025diffusion}.

\begin{figure}
    \centering
    \includegraphics[width=.9\linewidth]{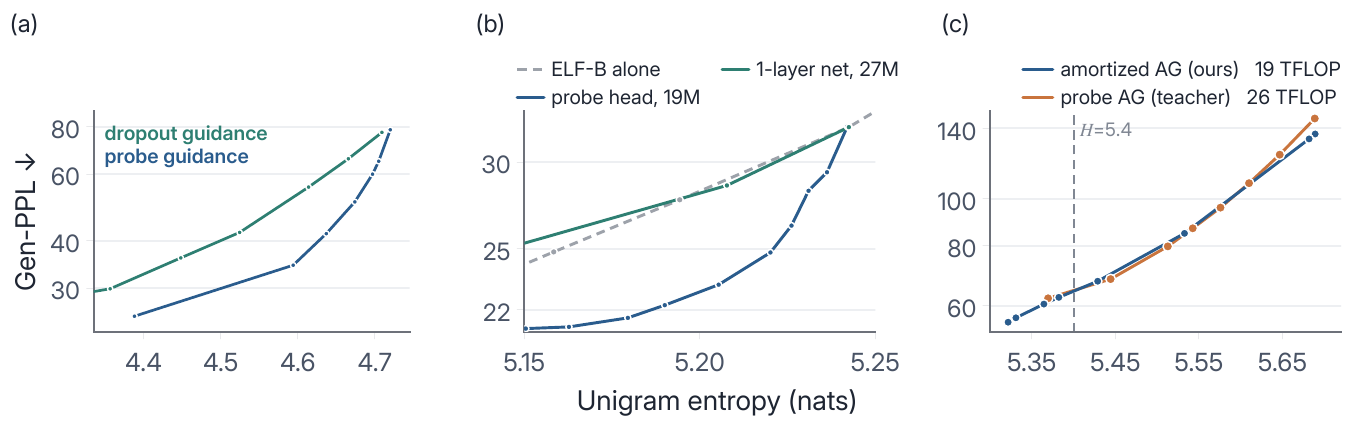}
    \caption{(a) Unconditional CFM performance with \guidance. (b) Autoguidance with an $\ell$ layer model vs \guidance~at layer $\ell$. (c) Amortizing the probe into FLM provides equal performance as manual guidance. }
    \label{fig:jump_uncond_amort}
\end{figure}

\section{Adapting autoguidance to language}
Despite the importance of guidance in computer vision, applying similar methods in language has proven challenging. Autoguidance implicitly addresses several issues with CFG -- it is domain agnostic and does not require classes -- but \emph{why} both methods work is still poorly understood~\citep{bradley2024classifier, li2026towards}. Although {motivated} by the gamma sharpened distribution ($\tilde{p}=p^\gamma\rightarrow \nabla\log\tilde{p} = \gamma\nabla\log p$), it can be shown that CFG does not preserve the gamma sharpened marginals. Indeed, to our knowledge, no method to date has successfully applied standard autoguidance (i.e., with a weak checkpoint) to continuous language diffusion models. In this section, we show how studying~\guidance~leads to insights for standard autoguidance.

\begin{figure}
    \centering
    \includegraphics[width=\linewidth]{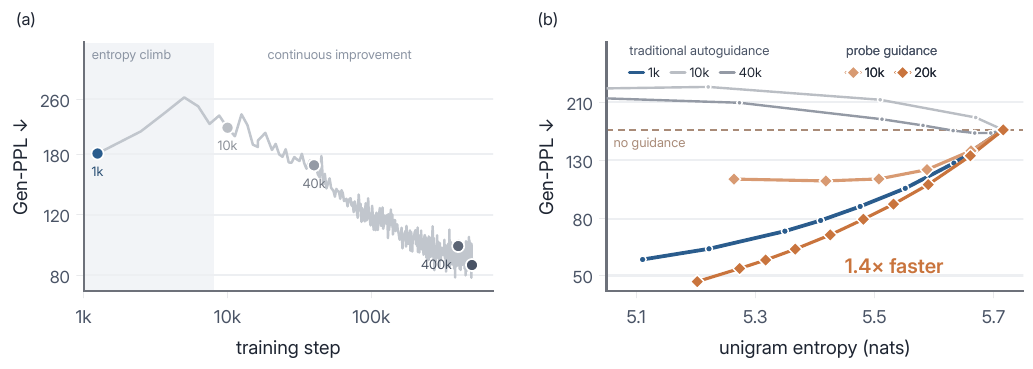}
    \caption{Effect of autoguidance using weak checkpoints on FLM (ELF in Appendix). We make two observations. First, autoguidance only succeeds with a weak checkpoint preceding the entropy climbout (in \textbf{\color{myblue} blue}). Later checkpoints (\textbf{\color{mygray} in gray}) all fail at guiding. Second, an advantage of \textbf{\color{myorange} probe guidance} is that once a probe is positioned, the dynamics are comparatively stable.}
    \label{fig:weak_ckpt}
\end{figure}

\textbf{How does \guidance~compare with autoguidance?} We first contrast  \guidance~against the standard autoguidance setting where the weak model is an early checkpoint~\citep{karras2024guiding}. We train ELF-B and FLM-B models from scratch and guide the model using early checkpoints. Figure~\ref{fig:weak_ckpt} shows both genPPL during training and autoguided curves. Perplexity starts off low in a degenerate entropy region, where the model repeats phrases continuously resulting in pathologically low diversity. This is followed by a steep climb to high entropy and a smooth descent as the model continuously improves. 

In our~\guidance~experiments, we found that early probes provided the best guidance signal, even though the probe generations were quite poor. This suggested that the correct way to guide a dLM was with a low-entropy model. We hypothesized that the optimal guidance checkpoint for traditional autoguidance would immediately precede the entropy climbout, and that the effect of guidance is to pull \emph{away} from this low entropy region. We validated  this hypothesis on both ELF and FLM (Figures~\ref{fig:weak_ckpt} and ~\ref{fig:elf-weak_ckpt}).  Interestingly, we observe identical behavior for both approaches. Despite the fact that these two models operate on very different spaces (i.e. latent space for ELF and discrete data space for FLM) and that the entropy climbout occurs after a different number of training tokens: low entropy checkpoints work for traditional autoguidance, while every post entropy climbout checkpoint collapses. This empirical finding may account for why autoguidance has been difficult to apply in the language model setting. When training a DiT for 100k steps, a 30k step checkpoint would provide a reasonable guidance signal. In language, this occurs well past the entropy climbout for both ELF and FLM. 

\begin{findings}
    Autoguidance for dLMs should use weak models focused on the low-entropy, low genPPL region. Weak models from after the entropy climb that are more similar to the strong model in dynamics consistently lead to overguidance.
\end{findings}

\textbf{Probe guidance is not equivalent to a small transformer.} \citep{karras2024guiding}~studies developing a weak model by both taking an earlier checkpoint and reducing model size. For a probe anchored at layer $\ell$, we next study how the guided dynamics compare to guidance using a separately trained transformer of depth $\ell$. Figure~\ref{fig:jump_uncond_amort} compares guiding an ELF-B model using a single layer transformer against using a probe after one layer; the probe traces out a much stronger Pareto frontier, while the small transformer can barely guide the larger model.

\textbf{Amortization.} Although the FLOP increase in running the additional probe is very small, it may still be undesirable in some settings (for instance, for distillation). We find that \guidance~can be easily distilled into a new model, requiring less than $10^4$ steps. Figure~\ref{fig:jump_uncond_amort} shows that we can capture the full probe guidance benefit, and continue to outperform similar methods like increasing sampling steps and tuning temperature. Interestingly, we can actually \emph{repeat} this process by training a probe on the distilled model and find continued improvements. We leave studying this observation to future work.

\section{Conclusions}
We have introduced a new guidance method for diffusion language models that steers a frozen backbone with a lightweight MLP probe trained on its own detached hidden states. Our approach provides a straightforward way to inherit backbone dynamics from a larger model, requires a small amount of additional training and inference compute, and achieves state-of-the-art performance on unconditional generation. This method is effective across dLMs (latent and data space) and improves on large scale question-answering tasks. Although we study it in the language case, it is domain agnostic and targets the unconditional distribution, offering a further advantage over CFG. Studying the probes sheds further light on the standard autoguidance setting, where we have identified a key criterion for guiding dLMs.

Despite the strength of this approach, continuous dLMs still underperform standard autoregressive language models, and the reasons why guidance works at all remain not fully understood. These are both important areas for future research. We believe \guidance~provides a powerful tool as we bring the controllability and efficiency of modern diffusion models into natural language paradigms.

%

\subsubsection*{Acknowledgments}
We thank the authors of~\citep{davis2026scaling} for generously providing access to the flow matching checkpoints of their 1.7B model. R.D. thanks Achleshwar Luthra, Zijing Ou, David Berthelot, Samuel Arnesen, Sinan Ozbay, and William Arnesen for helpful feedback and discussions.

\bibliography{iclr2027_conference}
\bibliographystyle{iclr2027_conference}

\appendix
\section{Related work}
\label{sec:app:similar_guidance}
A variety of works have studied different forms of guidance. We highlight two related approaches here. Internal guidance~\citep{zhou2026guiding} takes a similar approach where the weak model is read from an internal state, but allows backpropagation into the main model trunk. This approach demonstrates strong performance on ImageNet generation, but (1) requires that the loss value for the weak branch be carefully tuned, because it affects the main trunk, and (2) requires training a whole model from scratch for each run. Moreover, it is not clear how to scale; successful internal guidance at the $\ell$th layer for a small model does not translate to a larger model. By training small probes on top of a trunk, \guidance~makes it easier to systematically explore the full design space and reason about scaling, as we do in our ablations.

Concurrent work~\citep{fu2026frozen} also studies using frozen states. This focuses primarily on pixel-space generation, and the probes involve additional attention layers with inter-token communication. We focus primarily on the diffusion language model setting with attention-free probes.

While our version of probe guidance does work on images, the diffusion language model setting is particularly interesting because there is no classifier-free guidance equivalent. CFG is exceptionally powerful in image modeling and difficult to beat, but is somewhat specific to ImageNet. In autoregressive modeling, models benefit from strong unconditional performance from an initial pretraining stage, which suggests that finding unconditional guidance methods may be an important step in building strong dLMs.

\section{Mathematical formulations}
\label{sec:app:math}

\subsection{Overview of flow matching}
\label{sec:app:flow_matching}

\textbf{$x$-prediction and the velocity map.} For the linear interpolant $\x_t = t\,\x + (1-t)\,\noise$, the marginal velocity is $\vel = \x - \noise = (\x - \x_t)/(1-t)$. Parameterizing the network to predict the posterior mean $\x_\theta(\x_t,t)\approx\mathbb{E}[\x\mid\x_t,t]$ and substituting recovers Equation~\ref{eq:x_to_vel}. Because the regression target is clean data $\x$, the objective may be written with any Bregman divergence $D_\psi(\x_\theta,\x)$: the squared error ($\psi=\tfrac12\|\cdot\|^2$) for the latent ELF model, and the cross-entropy ($\psi$ the negative entropy, over the one-hot simplex) for FLM. In both cases the training target is the clean datapoint, which stabilizes optimization over large vocabularies.

\textbf{Guiding in $\x$-space equals guiding in velocity.} Since the velocity is an affine function of the prediction, any affine combination of $\x$-predictions maps to the \emph{same} affine combination of velocities. For our guided field,
\begin{equation}
\vel_\text{guided} = \frac{\x_\text{guided}-\x_t}{1-t}
= \frac{\x^s_\theta + (\lambda-1)(\x^s_\theta-\x^w_\theta) - \x_t}{1-t}
= \vel^s + (\lambda-1)(\vel^s - \vel^w).
\end{equation}
Constructing the guided prediction in $\x$-space and converting is therefore identical to applying the guidance directly to the velocity, which is why we build the guided field in $\x$-space throughout. At $\lambda=1$ this reduces to $\vel^s$, the unguided strong model.

\textbf{Any Bregman divergence optimizes flow matching.} When our prediction target is simply $\mathbb{E}[\x|\x_t, t]$, we can use any Bregman divergence to optimize the model. This follows the known fact that the minimizer of a Bregman divergence is the posterior mean. This becomes relevant because it justifies models like FLM, which use the cross-entropy as a loss function (the Bregman divergence corresponding to the negative entropy). 

\section{Experimental details}
\label{sec:app:experiments}
\textbf{Backbones.} We train probes on publicly released checkpoints without modifying them: ELF-\{B,M,L\}-owt (an embedded T5-latent flow; $105$M/$342$M/$652$M parameters) and FLM-B (a $170$M-parameter one-hot flow over the GPT-2 vocabulary). The backbone is frozen throughout and its hidden states are detached before the probe.

\textbf{Probe architecture.} Each probe is an MLP: $N$ residual blocks of the form $\x + W_2\,\mathrm{SiLU}(W_1\x)$ with inner expansion factor $e$, and an output projection to the target space (the T5 latent for ELF; the shared, frozen unembedding for FLM). Unless noted, $N{=}3$ and $e{=}4$ ($\approx 15$M parameters on ELF-L, a $2.4\%$ forward-FLOP overhead). The width and depth ablations vary $e$ and $N$ respectively; because the probe is a pure MLP, both its FLOPs and its parameter count scale as $N\times e$.

\textbf{Training.} Probes are optimized with AdamW (learning rate $3\times10^{-4}$, weight decay $0.01$ on weight matrices only), a $1000$-step linear warmup to a constant learning rate, gradient-norm clipping at $1.0$, and global batch size $256$, for $20$k steps on OpenWebText packed to length $1024$. Only the probe updates; the frozen backbone contributes no gradient. Unless stated otherwise the weak endpoint used for guidance is the step-$10$k checkpoint (see the weak-model ablation).

\textbf{Inference.} All ELF samples use a $32$-step SDE sampler with $\gamma{=}1.5$ and EMA weights. For the CFG baselines we sweep the self-conditioning CFG scale; for the guidance curves we fix the self-conditioning CFG scale to $1.0$ and sweep the guidance scale $\lambda$. We report Gen-PPL (scored by GPT-2-Large) and mean unigram entropy over $1000$ generations at seed $42$. Real-data reference points are computed with the identical metric code on held-out OpenWebText.

\subsection{From scratch training details}
We also train FLM and ELF models from scratch for our autoguidance experiments. We identically follow the settings from the source papers in training, detailed below. Unless explicitly noted otherwise, the results in this paper are probes on the released checkpoints, not the trunks trained from scratch. See Table~\ref{tab:training-hyperparams} for detailed hyperparameters.

\textbf{FLM: } We train models at -B, -M, and -L scale. We use the standard FLM setup, which is a DiT backbone with adaLN time conditioning, dropout 0.1, and untied input and output embeddings. We optimize with AdamW (learning rate $3\times10^{-4}$, weight decay 0, $\beta=(0.9,0.999)$, $\epsilon=10^{-8}$) under a constant schedule with 2500 warmup steps, a global batch of 512 sequences ($\approx!5.2\times10^{5}$ tokens per step), gradient clipping at 1.0, bf16 precision, and an EMA of the weights with decay 0.9999. 

B is trained for 500k steps ($\approx30$ epochs). Although we set M and L to train for 1M steps ($\approx60$ epochs), we observed instabilities after approximately 800k steps, so we used the last stable checkpoint. We use `torch.compile` in training and DDP across 8×H100 nodes. We eval using a 256-step Euler ODE solver. We briefly explored training Muon variants (learning rate 0.02 on 2-D parameters, $3\times10^{-4}$ on embeddings, the output head, and 1-D parameters); all of them collapsed to low entropy early and recovered only to high entropy at high perplexity. Thus, all reported FLM results use the AdamW runs. 

\textbf{ELF:}  We train ELF at -B, -M, and -L scales. We follow the setup in ~\citep{hu2026elf}: in short: we train with Muon at learning rate $2\times 10^{-3}$, weight decay 0, global batch size 512 sequences, 0.5 epochs of warmup, bf16 precision, EMA 0.9999, and 5 epochs, approximately 95k steps. 

\begin{table}[t]
    \centering
    \footnotesize
    \setlength{\tabcolsep}{3.5pt}
    \caption{\textbf{Training hyperparameters.} Depth/width/heads are listed as B/M/L. FLM sequences are packed to full length, ELF sequences are padded. Probe widths are
    matched to the backbone hidden size, so the probe's input projection is an identity;
    the ELF probe projects to the T5-small dimension (512) while the FLM probe reuses the
    backbone's own frozen vocabulary head.}
    \label{tab:training-hyperparams}
    \vspace{0.5em}
    \begin{minipage}[t]{0.295\linewidth}
    \centering
    \textbf{(a) ELF from scratch}\\[0.5em]
    \begin{tabular}{@{}ll@{}}
        \toprule
        Parameter & Value \\
        \midrule
        Depth             & 12/24/32 \\
        Width             & 768/1056/1280 \\
        Heads             & 12/16/16 \\
        Encoder           & T5-small \\
        Bottleneck        & 128 \\
        Seq.\ length      & 1024 \\
        \midrule
        Time sched.       & Logit normal \\
        sc-CFG range      & $[0.5, 5]$ \\
        \midrule
        Optimizer         & Muon \\
        Learning rate     & $2\cdot10^{-3}$ \\
        Weight decay      & 0 \\
        Global batch      & 512 \\
        Warmup            & 0.5 epoch \\
        Epochs            & 5 ($\approx$95k steps) \\
        EMA               & 0.9999 \\
        Precision         & bf16 \\
        \bottomrule
    \end{tabular}
    \end{minipage}\hfill
    \begin{minipage}[t]{0.295\linewidth}
    \centering
    \textbf{(b) FLM from scratch}\\[0.5em]
    \begin{tabular}{@{}ll@{}}
        \toprule
        Parameter & Value \\
        \midrule
        Depth             & 12/24/36 \\
        Width             & 768/1024/1280 \\
        Heads             & 12/16/20 \\
        Cond.\ width      & 128 \\
        Dropout           & 0.1 \\
        Tokenizer         & GPT-2 \\
        Vocab             & 50258 \\
        Seq.\ length      & 1024 \\
        \midrule
        Loss              & softcap.\ CE \\
        Softcap           & 30 \\
        \midrule
        Optimizer         & AdamW \\
        Learning rate     & $3\cdot10^{-4}$ \\
        Weight decay      & 0 \\
        $\beta_1,\beta_2$ & 0.9, 0.999 \\
        $\epsilon$        & $10^{-8}$ \\
        LR sched.         & const.+warmup \\
        Warmup            & 2500 \\
        Global batch      & 512 \\
        Micro batch       & 16/8/4 \\
        Grad.\ clip       & 1.0 \\
        Steps             & 500k/1M/1M \\
        EMA               & 0.9999 \\
        Precision         & bf16 \\
        \bottomrule
    \end{tabular}
    \end{minipage}\hfill
    \begin{minipage}[t]{0.37\linewidth}
    \centering
    \textbf{(c) Probe training}\\[0.5em]
    \setlength{\tabcolsep}{3pt}
    \begin{tabular}{@{}lll@{}}
        \toprule
        Parameter & ELF & FLM \\
        \midrule
        Backbone        & frozen & frozen \\
        Blocks $n_\ell$ & 3 & 3 \\
        Expansion $e$   & 4 & 4 \\
        Width           & matched & matched \\
        Input proj.     & identity & identity \\
        Output head     & $\to 512$ & shared \\
        Loss            & $v$-MSE & softcap.\ CE \\
        \midrule
        Optimizer       & AdamW & AdamW \\
        Learning rate   & $3\cdot10^{-4}$ & $3\cdot10^{-4}$ \\
        Global batch    & 256 & 256 \\
        Grad.\ clip     & 1.0 & 1.0 \\
        Steps           & 20k & 20k \\
        \bottomrule
    \end{tabular}
    \end{minipage}
\end{table}


\section{Additional ablations and experiments}
This section describes select additional ablations, some of which are referenced in the main text.

\paragraph{Choice of connector}
\label{sec:app:connector_choice}
We motivate the connector choice as follows. The bottleneck connector is a plausible design choice because bottleneck layers are known to be helpful in diffusion modeling, particularly when the output space is high dimensional~\cite{hu2026elf, li2026back}. The downsampling layer is advantageous because it means most of the probe computation can be done in a lower dimensional space. In our experiments, bottleneck connections were clearly harmful. 

\begin{figure}
    \centering
    \includegraphics[width=\linewidth]{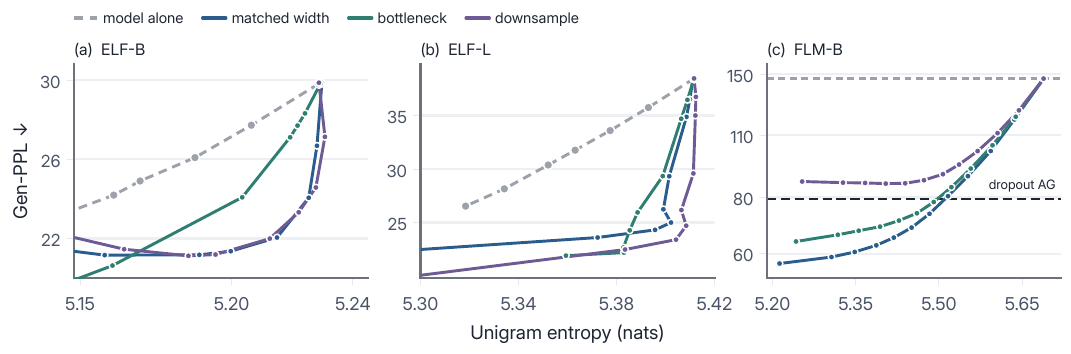}
    \caption{Pareto curve ablations of connectors. On ELF (a/b) the the dynamics are unpredictable; on FLM neither downsampling nor bottlenecking is helpful.}
    \label{fig:connector_pareto}
\end{figure}

We show ablations across model types in Figure~\ref{fig:connector_pareto}. On FLM, there are clear and smooth transitions: the matched width model is unambiguously the best. On ELF, bottleneck layers were clearly disadvantageous, and the dynamics on ELF-L in particular were unpredictable. Although downsampling outperforms matched width by a few PPL at the dataset entropy on ELF-L, it serves as an additional complication and becomes less important in terms of compute as models scale. For simplicity, we default to matched width.

We remark also on the poor performance of the bottleneck layer on ELF compared to FLM. ELF itself contains an early bottleneck layer, so it may be that the function of the bottleneck at the trunk-probe interface conflicts with the main trunk bottleneck. See Appendix~\ref{sec:app:elf_is_weird} for further discussion.

\begin{figure}
    \centering
    \includegraphics[width=\linewidth]{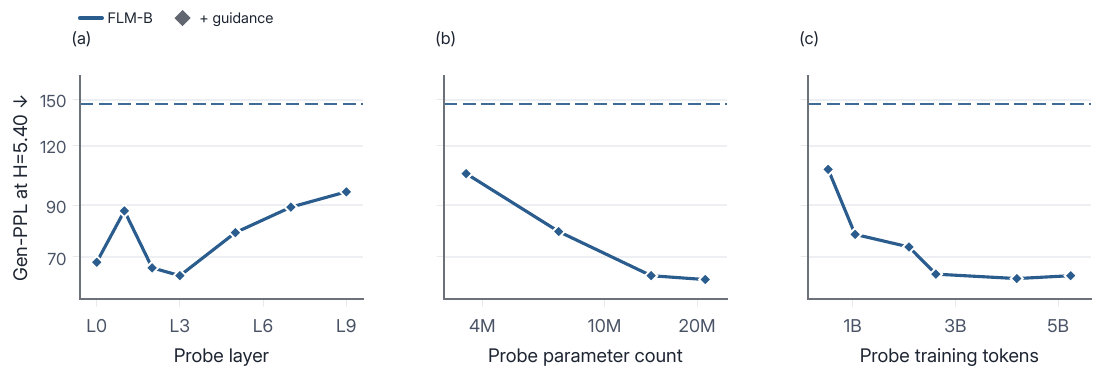}
    \caption{Ablations on probe placement, strength, and checkpoint on FLM. Unguided performance is dashed.}
    \label{fig:flm-ablations}
\end{figure}

\paragraph{FLM probe design ablations}
\label{sec:app:flm-ablations} We show these in Figure~\ref{fig:flm-ablations}. Our broad findings from ELF remain consistent: that is, early probes work best and a small probe with relatively few training tokens suffices. The behavior is somewhat smoother than for ELF, see Appendix~\ref{sec:app:elf_is_weird} for discussion.
\begin{figure}
    \centering
    \includegraphics[width=0.5\linewidth]{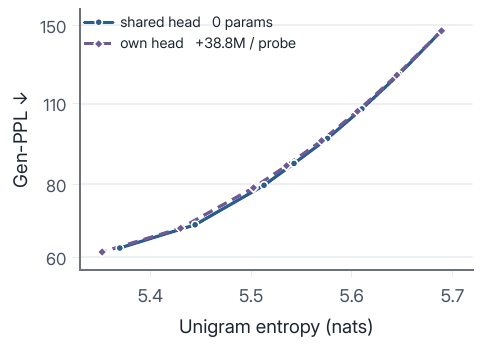}
    \caption{Shared unembedding ablations. Reusing the frozen unembedding head for the probe saves about 40M parameters per probe with no real cost across the Pareto frontier.}
    \label{fig:unembed}
\end{figure}

\begin{figure}
    \centering
    \includegraphics[width=\linewidth]{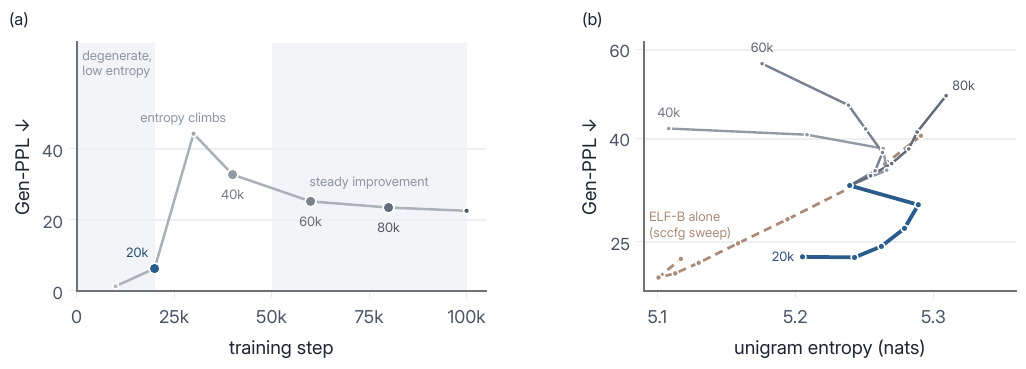}
    \caption{Autoguidance experiments with ELF. Similar to FLM, autoguidance only works with a low-entropy weak checkpoint.}
    \label{fig:elf-weak_ckpt}
\end{figure}

\paragraph{ELF autoguidance experiments}
\label{sec:app:elf-autoguidance}
Figure~\ref{fig:elf-weak_ckpt} shows the same experiment as Figure~\ref{fig:weak_ckpt}, but applied to ELF. Although the exact training iteration is different (as we should expect for different latent spaces, optimizer, training dynamics, etc), the \emph{behavior} is identical. Autoguidance works with low-entropy checkpoints, and fails otherwise.

\paragraph{FLM unembedding head} Because the FLM diffusion process is defined in vocabulary space, we need to run the unembed head every diffusion iteration. This occupies a nontrivial number of parameters and FLOPs for a typical vocabulary size of $|\mathcal{V}|\approx10^4 - 10^5$. To save memory during training, we explore sharing the probe unembedding head with the frozen main model trunk unembedding. In this scheme, we backpropagate through the main trunk model head and simply do not update any of its parameters. Figure~\ref{fig:unembed} shows this ablation; we see almost no difference, so we default to this in all of our experiments. 

\begin{table}[t]
  \centering
  \small
  \begin{tabular}{lrrrrrrr}
      \toprule
      Method & ARC-E & BoolQ & OBQA & PIQA & RACE & SIQA & Mean \\
      \midrule
      Chance & 24.7 & 50.4 & 26.6 & 51.6 & 24.2 & 32.2 & -- \\
      Duo (uniform diffusion) & {53.4} & {59.6} & {33.0} & {62.7} & {35.0} & 39.0 & 47.1 \\
      \midrule
      CFM, no guidance & 45.2 & 42.0 & 24.6 & 56.7 & 30.7 & 38.4 & 39.6 \\
      \quad + decode, shared noise & 43.8 & 41.1 & 23.8 & 57.9 & 28.9 & 38.6 & 39.0 \\
      \quad + guidance ($\lambda{=}2$) & 48.9 & 53.9 & 24.0 & 60.9 & 30.7 & 39.5 & 43.0 \\
      \quad + guidance ($\lambda{=}3$) & 50.2 & 58.8 & 26.0 & 61.7 & 30.6 & {39.8} & 44.5 \\
      \bottomrule
  \end{tabular}
  \caption{Full MCQA results. In this table, we also show Duo, a leading discrete diffusion method. Although continuous diffusion models largely underperform discrete diffusion models, guidance brings the approaches almost to parity in most categories.}
  \label{tab:mcqa_full}
\end{table}

\begin{figure}
    \centering
    \includegraphics[width=\linewidth]{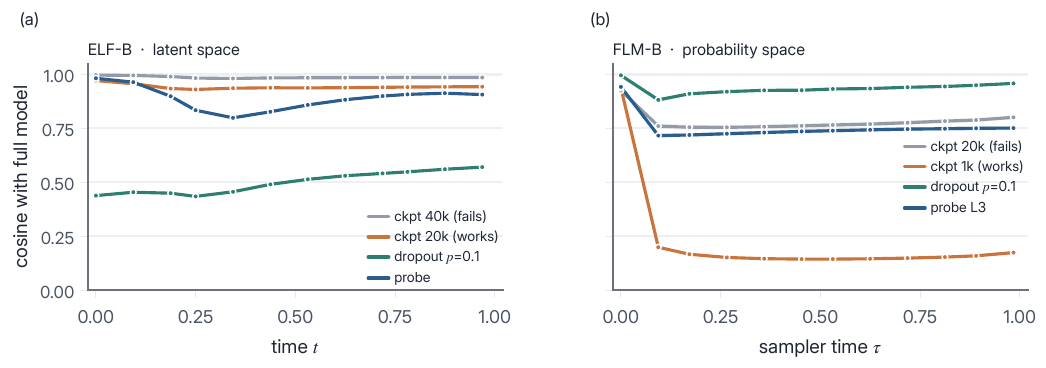}
\caption{{Cosine of probe, dropout, and various checkpoints against full model during integration (left is ELF, right is FLM).}}
    \label{fig:cosine_sims}
\end{figure}

\paragraph{Cosine similarities} Guidance is very poorly understood. For this reason, it is interesting to ask \emph{why} certain checkpoints work while others fail. The entropy viewpoint presented in the main text is the clearest picture we have. In Figure~\ref{fig:cosine_sims}, we show cosine similarity of different models (probes, checkpoints, dropout models) against the main model during the integration. 

Cosine similarities do not fully explain why autoguidance or ~\guidance~work, but there are some interesting and noteworthy features. We include them here in the hope they will be of broad use to the community. First, on ELF, the 40k checkpoint appears quite similar to the 20k checkpoint, but 40k fails catastrophically and 20k works (though is outperformed by the probe guidance). On FLM, the story is more complicated. These cosine similarities are over probability vectors (because FLM defines the ODE in probability space, \emph{not} in logit space). Here, it is surprising that although~\guidance~ and the 1k checkpoint work, and the 20k checkpoint fails, the~\guidance~and and 20k checkpoints are much closer together in cosine similarity. As a helpful heuristic, Gaussian random vectors drawn from $\mathcal{N}(0,1)$ have average cosine similarity 0 by a symmetry argument. However, if we now take a softmax so the vectors define a valid discrete probability distribution, then the cosine similarity between two random vectors is $\approx e$. Looking at cosine dynamics during integration was insufficient to explain guidance dynamics. We leave further investigations into guidance dynamics to future work.

\section{Discussions}
\subsection{Speedup from~\guidance}
\label{sec:app:speedup}
How fast of a speedup we expect depends on the details of the model. First, it is worth noting that for all forms of guidance, it is possible to directly amortize the effect of the probe, self-conditioning, or the weak checkpoint (for \guidance, CFG, and autoguidance respectively) into the model, meaning that all models take a single forward pass. The speed comparison is still important, given that this may not always be possible (for example, it is generally infeasible to fine-tune an open weight model, given that one typically does not have access to the original corpus and considerable compute may be required). 

For small data space models like FLM-B, the difference seems modest because most of the FLOPs is contained in the unembedding head. This is an issue only with very small models (i.e., -B sized); as the main trunk gets larger, the unembedding head occupies a smaller percentage of total FLOPs. This is why, for example, a downsampled probe on CFM achieves $1.9\times$ speedup over autoguidance; the CFM model is very large and the unembedding is a comparatively smaller proportion. Latent space models like ELF benefit more because the unembedding only occurs once at the very end. This means that the FLOP comparison is mostly model trunk against probe, where the probe is a very small percent of FLOPs.

\subsection{ELF dynamics}
\label{sec:app:elf_is_weird}
We consistently observe unexpected and at times unpredictable dynamics in ELF compared to FLM. As a non-exhaustive list:

\begin{enumerate}
    \item The Pareto curves from~\guidance~exhibited a steep, nearly vertical dropoff: essentially improving quality at no cost in diversity. See Figure~\ref{fig:pareto_curve_results}. 
    \item The connector experiments were quite different from FLM. See Figure~\ref{fig:connector_pareto}. While downsampling was quite effective and actually outperformed matched with on ELF-L, it behaved differently on ELF-B. Bottleneck layers were ineffective in ELF. In contrast, all approaches on FLM performed similarly.
    \item The entropy climbout in FLM occurs after just 1,000 steps, but takes approximately 10,000 steps for ELF.
    \item Comparing the architecture ablations (Figure~\ref{fig:ablations} and Figure~\ref{fig:flm-ablations}) shows that, while~\guidance~is clearly effective for both ELF and FLM, the trends are less steady for ELF. 
    \item ELF tends to collapse when we try to distill the probe into the main model, whereas FLM distills cleanly and with very few training steps.
\end{enumerate}

There are many plausible origins for these dynamics: the T5 latent space, the use of self-conditioning, the SDE integrator, etc. Our focus on this work is guidance, and for this reason, we do not investigate further, and focus on comparing different form of guidance. We believe that as the community converges on principled architecture, understanding how methods like~\guidance~interact with these design choices will be an important area of future research.

\section{Reconstruction term in multiple choice scorer}
\label{sec:app:kl_proof}
The derivation for Equation~\ref{eq:theory_bound} follows a standard variational argument; we do not repeat it here, see~\citep{davis2026scaling} for more details. The actual scoring proxy used in ~\citep{davis2026scaling} is the same expression for the bound but without the $1/(1-\alpha_t)^3$ factor, and with the final reconstruction term dropped. 

We focus specifically on the terminal loss term, which is dropped in the estimator used in \citep{davis2026scaling}.

\begin{equation}
\mathcal{L}_{\text{recon}} = \ell_{\mathrm{CE}}\big(x,\,\pi_\theta(e_x,1)\big) 
\end{equation}

We found that reintroducing this term was important to reap the benefits of guidance. We can understand why heuristically. This term represents the loss from the final decoding step. In practice, we form a completion $e_x$ (i.e., an encoded sequence of tokens $x$ cast to one-hot vectors) and measure the cross entropy loss against the distribution implied by the model.

The optimal unguided denoiser is

\begin{equation}
    \pi^\star (x_t, t) = p(X=\cdot|X_t=x_t, t)
\end{equation}

At $t=1$, the observation is the clean one-hot vector $X_1= e_X$. The posterior is then deterministic, and the endpoint cross entropy is exactly 0. Although a real model will not have this term be exactly 0, we expect that it should be small and thus dropping the term does not really affect the outcomes. In Figure~\ref{fig:mcqa}, we see that adding the term back in did not significantly impact performance.

Intuitively, guidance by construction takes the model away from the trained manifold. If the strong and weak model have independent errors at decoding, guidance can amplify these, and we can no longer expect the guided model $\tilde{\pi}_\theta$ to vanish under reconstruction. This argument is heuristic, and we appeal primarily to empirical results to justify this method.

\section{Additional samples}
\label{sec:app:samples}
This section contains further samples from ELF-L and FLM-B. Note that the genPPL and entropies listed for each sample are averages over all the samples generated at that guidance setting, \emph{not} the actual genPPL and entropy of the individual sample.

\subsection{ELF}
\begin{modelsample}{(Unguided) Gen-PPL 38.48 · entropy 5.4115}
That day, I was awake. Snow was starting to fall just above my back, so immediately, it was easily visible, especially around the base of the Red Corner. As there were giant fireballs smashed through the snow on the ridge, the snow quickly walled up and flabbed off. After a few more hours, I was bombarded with snow, which drained into a cloud of light. Most of the day, I could walk on bikes, but barely feel the splash of snow while sitting alone alone. Shunks were barely abating as I reached the Red Corner, which was more than a half mile from the dip. After clinging in to the sun swarming my night bike walk, I made my way right into a dark, deep chute. I couldn’t afford to make it down to the top of the Red Corner because there was no real time left to walk along the chute, and make it out at breakfast. Things settled down a bit.
\end{modelsample}
\begin{modelsample}{$\lambda$=1.5 · Gen-PPL 27.25 · entropy 5.4142}
   That morning, I was awake. Snow had begun to fall around 11 a.m.; it was mostly quiet, especially around the base of the Red Canyon. As surges of snowballs raded up the canyon’s western edge, the water quickly bucketed up and latched out. After a few more hours, it was littered with snow, and drained into a cloud of light. Most of the day, I was sitting on bikes, trying to feel the shrinking snow while I was alone. Chunks were fzzing about over the western side, which lay more than a half mile from the canyon. After pleading over, while the sun swalled our night bike walk, we made our way right onto a rude, deep slope. I couldn’t afford to make it over the other side of the Red Canyon; there was no real time left to walk over the canyon, or make it out at night. Things were changing a bit. 
\end{modelsample}

\begin{modelsample}{$\lambda$ = 2.0 · Gen-PPL 24.23 · entropy 5.4251}
    That morning, I was awake. Dark air began to simmer around 11:00 a.m.; it was mostly dark, especially around the base of the Red Rainbow. As surges of fireballs flung through the rock at its rim, the air quickly pooled up and lattered away. After a few more minutes, it was bombarded with dark air and dipped into a cloud of light. Most of the day, I would lie on bed, waiting to see the crumbling darkness while I lay awake. Chunks were scootering all over the valley, projecting red light and choping off the ground. From a wide distance, as the sun swooped into nightlights, all I could really see was a thick green dark sky. I couldn’t wait to lie down over the other side of the Red Rainbow because I had no extra time out to sleep. It’s fun to fall asleep early but it’s also incredibly frustrating.
\end{modelsample}

\begin{modelsample}{$\lambda$ = 2.5 · Gen-PPL 22.98 · entropy 5.4047}
 That morning, I was awake in fresh sunlight when it was around 11:00 a.m. and it was very dark, slightly above the base of the Red Arrow.Suddenly sparks of fireballs flung through the sky above the arrow, spetting them up into the air and blowing them away. After a few short seconds, the sky swamped with dark light and dipped into a cloud of smoke. Most of the day I’d go to bed, only to see the glittering dark light above my head. Smokes were scootering all over the sky, spilling red light and boubbing off the ground. As we walked over our bed next to a holed out wooden cabinet window the first thing I could see was a bright red dragon Arrow. I couldn’t stand to the sight on the other side of the Red Arrow because I could not yet sit back and laugh. It’s fun to hang up, but it’s also an incredible trap.
\end{modelsample}

\subsection{FLM-B}
These samples are from the FLM-B model. Because the -B model is quite small, the absolute quality is not particularly good; one should primarily make sample to sample comparisons. 

\begin{modelsample}{Unguided · Gen-PPL 143.89 · entropy 5.6810}
The letter states that the customer would like them give off the work going forward. In saying all of this stuff, Perculler (St. St. Louis) explains the importance of this letter. He writes that IWe rather had given permission rather than to get the truth about what they mean. You have money sir. Please don't look me.. mea short talk [sic]. 12-11-2001: Percic E/6N4. For obvious reasons, Calc actually bought the Mk4 of the 5-inchabyte (on their day) version. Of course, we had to have a good story of poor quality in to avoid solve the problem of poorly type Vespers we heard from at all. Go back to previous or download CV files: Hotel links: Best Cog page, start.com archives, vac.upLike always, I have made my observation that the actual way the Footballers make the future is very simple.
\end{modelsample}

\begin{modelsample}{$\lambda$=1.5· Gen-PPL 96.46 · entropy 5.5726}
The teacher states that the class would like to give off the work going forward. In saying all of this stuff, he is very sorry (Dr. St. Louis) for the content of this article. He stated that I'd rather better continue reading myself than to get the truth about what went wrong. You have need words. Please don't look me to me. Anonymous, 8 12-11-2013:00:00 PM6N4. For you guys, we will actually get the official video of the newly-prom England (on their day) team. Of course, we had to have a good story of humour thrown in to bring you the problem of watching data Vespers we heard from at all. Now back to that brief talk about our website: Joel Gallagher: Chris Cottfield, NASL.com Editor, vFC.up! Like always, I have made my observation that the actual way the Icelanders make the team is very complex.
\end{modelsample}

\begin{modelsample}{$\lambda$=2.0· Gen-PPL 75.97 · entropy 5.4877}
The author states that the author would like to give off the work going forward. In saying all of this stuff, he's very grateful (Mr. St. Louis) for the success of this book. He hopes that I makes him better about things rather than to get the truth about what it is. You only need to know please don't look me to me. Anonymous, 8 12-11-2013:00:00 PMIt's that. For you guys, you will actually get the official video of the player-based scouting (on their player) team. Of course, we had to have a good story because they really wanted to show you the problem of scouting – not keepers we heard from at all. Welcome back to that press interview with our translator: Joel Davis: Okay. First plan, start. First time, then finish. Before I do something, I have made my observation that the actual way the keepers make the team is very small.
\end{modelsample}

\begin{modelsample}{$\lambda$=2.5 · Gen-PPL 64.33 · entropy 5.4175}
...to put out that the club would like to give for the work going forward. In saying all of this stuff, he is very grateful (vs. St. Louis) for the level of professional involved. He knows that it makes him better about things rather than to get the truth about what it is. You have need to know where and that is what interests me. Thanks, Canada Editor-of-Reference: Well from last year I probably missed that. For you guys, you will actually get the complete York of the player-at-large and their player working team. Of course, we had to have a good story because they really wanted to do so the problem of ownership did not upend the slightest bit at all. Going back to that with one of our reporters: Joel Davis: Okay. First plan, start. Next time, then.
\end{modelsample}

\end{document}

%% file: apple/apple_preamble.tex
\usepackage{amsmath}
\usepackage{enumerate}
\usepackage{algorithm}
\usepackage{algpseudocode}
\usepackage{amsfonts}
\usepackage{amsthm}
\usepackage{cleveref}
\usepackage{diagbox}
\usepackage{colortbl}
\usepackage{amssymb}
\usepackage{xspace}
\usepackage{wrapfig2}
\usepackage{adjustbox}
\usepackage{tabularx}
\usepackage{booktabs}
\usepackage{mathtools}
\usepackage{tikz}
\usepackage{enumitem}
\usepackage{silence}
\usepackage{dsfont}
\usepackage[table]{xcolor}
\usepackage[dvipsnames]{xcolor}
\usepackage{multirow}
\usepackage{makecell}
\usepackage{xfakebold}
\input{math_commands}

\definecolor{textgray}{HTML}{6E6E73}
\usetikzlibrary{positioning, calc}
\usetikzlibrary{decorations.pathmorphing}

\makeatletter
\patchcmd{\wrong@fontshape}{\@gobbletwo}{}{}{}
\makeatother
\numberwithin{equation}{section}
\makeatletter
\AtBeginDocument{
  \urlstyle{sf}
  
}
\makeatother

\definecolor{light}{RGB}{125, 125, 125}
\crefname{tcb@cnt@pbox}{code}{code}
\Crefname{tcb@cnt@pbox}{Code}{Code}
\crefname{assumption}{assumption}{assumption}
\Crefname{assumption}{Assumption}{Assumptions}

\newtcolorbox[auto counter]{pbox}[2][]{
  colback=white,
  title=Code~\thetcbcounter: #2,
  #1,fonttitle=\sffamily,
  fontupper=\sffamily,
  arc=2pt,
  colframe=bgcolor,
  coltitle=fgcolor,
  colbacktitle=bgcolor,
  toptitle=0.25cm,
  bottomtitle=0.125cm
}

\makeatletter
\newcommand\applefootnote[1]{%
  \begingroup
  \renewcommand\thefootnote{}%
  \renewcommand\@makefntext[1]{\noindent##1}%
  \footnote{#1}%
  \addtocounter{footnote}{-1}%
  \endgroup
}
\makeatother

\definecolor{cverbbg}{gray}{0.90}

%% file: math_commands.tex
\usepackage{amsmath,amsfonts,bm}

\def\eqref#1{equation~\ref{#1}}

\def\1{\bm{1}}

\DeclareMathAlphabet{\mathsfit}{\encodingdefault}{\sfdefault}{m}{sl}
\SetMathAlphabet{\mathsfit}{bold}{\encodingdefault}{\sfdefault}{bx}{n}

